\documentclass[letterpaper, 10 pt, journal, twoside]{IEEEtran}
\ifCLASSINFOpdf
\else
\fi
\usepackage{mathtools}
\usepackage{amsmath}

\usepackage{amssymb}
\usepackage{amsfonts}
\usepackage{amsopn}
\usepackage{graphicx} 
\usepackage{textcomp}
\usepackage{xfrac}
\usepackage{bbm}
\usepackage{overpic}
\usepackage{subfig}
\usepackage{wrapfig}
\usepackage[ruled,vlined,linesnumbered]{algorithm2e}

\SetAlFnt{\small}
\SetAlCapFnt{\small}
\SetAlCapNameFnt{\small}
\SetAlCapHSkip{0pt}

\usepackage{multirow}

\usepackage{booktabs}
\usepackage{footnote}
\usepackage{paralist}
\usepackage{enumitem}
\usepackage{autobreak}
\usepackage{hyperref}
\usepackage{cleveref}
\usepackage[thicklines]{cancel}

\usepackage{color}
\definecolor{green}{rgb}{0, 0.4, 0}
\definecolor{orange}{rgb}{0.8, 0.6, 0.2}
\definecolor{red}{rgb}{1.0, 0.0, 0.0}
\definecolor{teal}{rgb}{0.0, 0.4, 0.4}
\definecolor{purple}{rgb}{0.65,0,0.65}
\definecolor{saffron}{rgb}{0.95,0.75,0.2}
\definecolor{turquoise}{rgb}{0.0,0.5,0.5}
\definecolor{brown}{rgb}{0.5, 0.16, 0.16}

\usepackage{overpic}
\usepackage{currfile} 

\newlength\savedwidth

\definecolor{lightgray}{rgb}{0.6, 0.6, 0.6}

\newcommand{\changed}[1]{{\color{black}#1}}

\usepackage[normalem]{ulem}

\newcommand{\hidecomment}[1]{}

\newcommand{\cF}{\mathcal{F}}
\newcommand{\cS}{\mathcal{S}}
\newcommand{\cP}{\mathcal{P}}

\newcommand{\cL}{\mathcal{L}}

\newcommand{\bn}{\mathbf{n}}
\newcommand{\br}{\mathbf{r}}

\newcommand{\bp}{\mathbf{p}}

\newcommand{\bphi}{\mathbf{\phi}}

\usepackage{xspace}

\begin{document}
%
\title{Neural Observation Field Guided Hybrid Optimization of Camera Placement}
%
%
%

\author{Yihan Cao$^{1}$, Jiazhao Zhang$^{2}$, Zhinan Yu$^{1}$, and Kai Xu$^{1, {\dagger}}$
\thanks{Manuscript received: March, 28, 2024; Revised June, 19, 2024; Accepted July, 26, 2024.}
\thanks{This paper was recommended for publication by Editor Pascal Vasseur upon evaluation of the Associate Editor and Reviewers' comments.

This work was supported in part by the NSFC (62325211, 62132021) and the Major Program of Xiangjiang Laboratory (23XJ01009).} 
\thanks{$^{1}$ National University of Defense Technology.}%
\thanks{$^{2}$ CFCS, School of Computer Science, Peking University.}
\thanks{$^{\dagger}$ Corresponding Author.}%
\thanks{Digital Object Identifier (DOI): see top of this page.}
}
%
%

\markboth{IEEE Robotics and Automation Letters. Preprint Version. Accepted August, 2024}
{Cao \MakeLowercase{\textit{et al.}}: NeOF Guided Hybrid Opt. of Camera Placement} 

%



\maketitle

\begin{abstract}


Camera placement is crutial in multi-camera systems such as virtual reality, autonomous driving, and
 high-quality reconstruction. The camera placement challenge lies in the nonlinear nature of high-dimensional parameters and the unavailability of gradients for target functions like coverage and visibility. Consequently, most existing methods tackle this challenge by leveraging non-gradient-based optimization methods.
 %
%
In this work, we present a hybrid camera placement optimization approach that incorporates both gradient-based and non-gradient-based optimization methods. This design allows our method to enjoy the advantages of smooth optimization convergence and robustness from gradient-based and non-gradient-based optimization, respectively. To bridge the two disparate optimization methods, we propose a neural observation field, which implicitly encodes the coverage and observation quality. The neural observation field provides the measurements of the camera observations and corresponding gradients without the assumption of target scenes, making our method applicable to diverse scenarios, including 2D planar shapes, 3D objects, and room-scale 3D scenes.
Extensive experiments on diverse datasets demonstrate that our method achieves state-of-the-art performance, while requiring only a fraction (8x less) of the typical computation time. Furthermore, we conducted a real-world experiment using a custom-built capture system, confirming the resilience of our approach to real-world environmental noise.
\changed{We provide code and data at: \href{https://github.com/yhanCao/NeOF-HybridCamOpt}{https://github.com/yhanCao/NeOF-HybridCamOpt}.}

\end{abstract}


%
\IEEEpeerreviewmaketitle

\section{INTRODUCTION}
\IEEEPARstart{C}{amera} placement is a long-standing and widely applicable problem~\cite{kritter2019,kang2015progressive} across various domains such as motion tracking~\cite{zhou2022,rahimian2016}, surveillance systems~\cite{ahmad2022}, and robotics~\cite{zhang2021}. Through analysis of 3D spatial priors, camera placement methods optimize placement of multi-cameras, to maximize visibility metrics such as coverage. 
However, the camera placement problem faces two primary challenges: firstly, a highly non-linear optimization landscape due to the high dimensionality of multi-camera parameters; secondly, the process of calculating visibility lacks differentiability, rendering the gradient of visibility unattainable.
\begin{figure}[t]
\centering
\begin{overpic}
[width=\linewidth]
{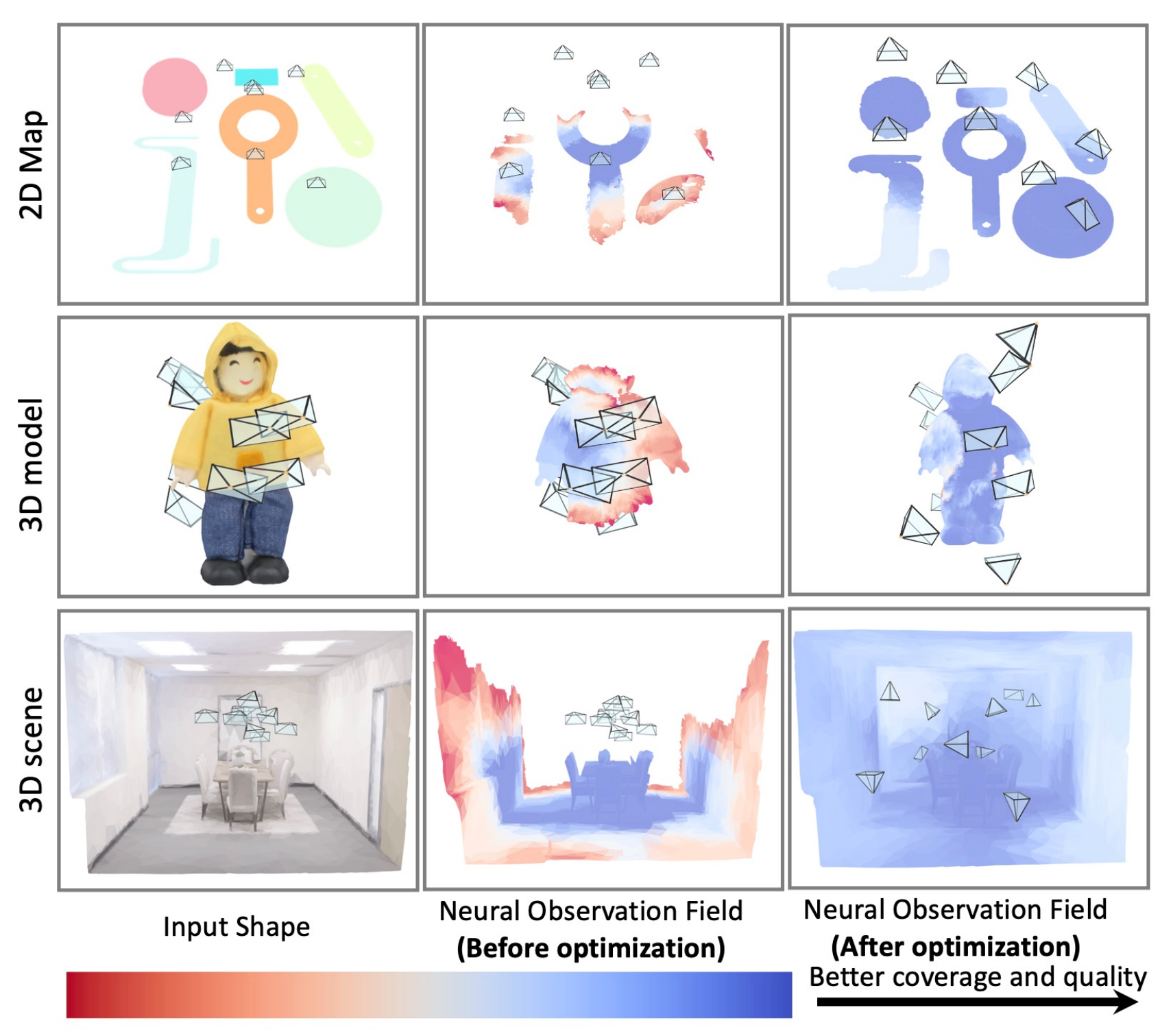}
\vspace{-2mm}
\end{overpic}

\caption{
We introduce a hybrid optimization method based on the neural observation field for camera placement estimation. The target objects are represented by neural observation fields, which are compatible with any type of objects.
}
\vspace{-6mm}
\label{fig:teaser}
\end{figure}

To tackle this challenge, most existing methods~\cite{ahmad2022,allah2022,malhotra2022,puligandla2023continuous} leverage non-gradient optimization methods with explicit scene representation, such as point clouds or voxels. These methods typically adopt heuristic or greedy
search algorithms, eliminating the need for gradients. However, they often suffer from issues related to convergence speed and precision. Additionally, explicit scene representations are predefined for specific scenarios, such as 2D planar shapes or 3D objects, leading to significant performance drops under different scenarios.



In this work, we propose a novel hybrid optimization approach for camera placement optimization, incorporating both gradient-based and non-gradient-based 
optimization methods. Fig.\ref{fig:teaser} illustrates the hybrid optimization results based on our proposed neural observation field. 

For gradient-based optimization, we propose a neural observation field designed to encode scene observation, facilitating gradient calculation for maximizing coverage and observation quality.~\changed{This neural observation field, represented by an implicit neural field, encodes the joint configuration of scene geometry with multi-camera placement, thereby implicitly capturing multi-view observability.} Throughout the optimization process, the neural observation field undergoes online updates following each new camera placement iteration. Leveraging the differentiability of the neural observation field, we compute gradients through gradient backward propagation and subsequently update camera placements. This approach ensures a smooth and rapid convergence of camera placement optimization.

For non-gradient-based optimization, we leverage an elite selection algorithm that retains cameras with high visibility while resampling those with poor visibility. The cameras with poor visibility are relocated to areas requiring more observation globally. This strategy reduces the risk of getting stuck in local optima and enhances the robustness of our method in complex scenarios, such as regions with large gaps or non-convex geometries.~\changed{Unlike approaches such as simulated annealing~\cite{shahrokhzadeh2017} or evolutionary strategies~\cite{rangel2019}, which involve resampling the placement of all cameras, our method focuses solely on resampling cameras with poor visibility, as identified by the convergent results of gradient-based optimization.} This approach is more efficient since cameras with good placement exhibit smaller gradients, which remain relatively consistent during gradient-based optimizations.


%


We conduct extensive experiments on diverse datasets, including 2D planar shapes, 3D objects~\cite{Downs2022GoogleSO}, and room-scale 3D  scenes~\cite{straub2019replica}. We evaluate our method on both coverage and observation quality. The results demonstrate that our approach outperforms existing solutions while requiring only a fraction (8x less) of the typical computational time. 
Furthermore, we develop a custom capture system to evaluate our method's resilience to real-world environmental noise. We find that our method demonstrates robustness and outperforms existing methods across a wide range of objects.

In summary, our main contributions include:

\begin{itemize}
    \item A hybrid camera placement optimization method that cooperatively incorporates both gradient-based and non-gradient-based optimization methods.
    \item A neural observation field that implicitly encodes the geometry priors and observation of indoor scenes.
    \item A custom-built capture system powered by our camera placement optimization method, demonstrating robustness and well-visibility in the real-world environments.
\end{itemize}

\section{RELATED WORKS}
\subsection{Camera Placement Application.}
Multi-camera systems are extensively employed for capturing and synthesizing realistic 3D content in both research and real-world development scenarios \cite{kritter2019}. Visual sensors, known for their low cost, lightweight, and image capture capabilities
in various domains. Multi-view applications have emerged in fields such as industrial inspection \cite{ gjakova2022}, surveillance \cite{zhang2021, ahmad2022}, motion capture \cite{rahimian2016}, and navigation \cite{zheng2019active,zhang2020fusion}, overcoming the limitations of single-camera vision.
\subsection{Camera Placement Optimization Methods.}
Multi-view optimization is a well-established NP-hard problem, making it impractical and time-consuming to find the optimal solution. The focus of this study is to find a sub-optimal solution within a reasonable time frame. Various approaches have been employed for this task \cite{zhang2019, bouhali2021}. Greedy algorithms, as utilized in \cite{brown2017}, iteratively consider all visibility factors until achieving high-quality results.
The computation time of the greedy algorithm escalates with scenario scope. 
Considering time overhead,~\cite{Navin2010SolvingCP, allah2022, shahrokhzadeh2017} optimize camera layouts using heuristic algorithms.
\cite{feng2021} employs Integer programming algorithm to optimize visibility, single-best coverage quality, and cumulative quality.~\cite{tao2023research} utilizes Particle Swarm Optimization (PSO) to minimize measurement error pixel quantization and measurement based on the environment model. However, the PSO algorithm's high computation time is noted in~\cite{puligandla2023continuous}.
While considering reconstruction, camera placement problem escalates from a single-coverage problem to a multi-coverage problem.
The placement of cameras also significantly influences the quality of the reconstructed 3D content.~\cite{rahimian2016} uses Simulated Annealing Algorithm (SA) to optimize camera placements by estimating the three-dimensional positions of markers through triangulation from multiple cameras.~\changed{For 3D reconstruction, bundle-based methods in \cite{olague2002automated,olague2002optimal,saadatseresht2004visibility} utilize photogrammetric camera networks or expert systems to obtain automated camera placement, considering only the reconstruction quality.}

The heuristic function for these tasks primarily revolves around the coverage rate of the entire scenario. Additionally,~\cite{malhotra2022} minimizes the coverage optimality gap, defined as the squared error between the desired and achieved coverage. This metric serves as our coverage metric for assessing multi-camera coverage.

Greedy-based or heuristic algorithms overlook the prior information regarding the shape of the scenario, resulting in time wastage in irrelevant areas. Moreover, the spatial discretization of solutions can lead to sub-optimal results, as it depends on the scale of spatial division. Our method incorporates both gradient-based and non-gradient-based optimization methods by taking advantage of scenario geometry prior to mitigate these challenges.
\subsection{Scene Representation for optimization.}
Scene representation in the camera placement problem involves unstructured point clouds~\cite{tao2023research}, volumes~\cite{malhotra2022}, or polygonal meshes~\cite{puligandla2023continuous}. 
However, model represented by points or meshes may degrade the result due to irregular distributions of points or faces,  which fail to adequately describe the coverage of model space, influenced by density~\cite{puligandla2022multiresolution}.
Voxelized scene representation, while effective in some cases, cannot facilitate gradient backpropagation for gradient-based optimization. Recently, implicit representation of shapes has made significant advancements. Implicit model field methods typically aim to learn a function that maps spatial locations to feature representations.
\cite{liu2020} proposes neural sparse voxel fields as a novel neural scene representation for fast and high-quality free-viewpoint rendering. In \cite{li2022}, a constructed voxel field representation is utilized for cross-modality 3D object detection.
Inspired by these works, we encode the scenario geometry prior to obtain the visibility of scenes.

\section{METHOD}
\begin{figure*}[t]
\centering
\begin{overpic}
[width=\linewidth]
{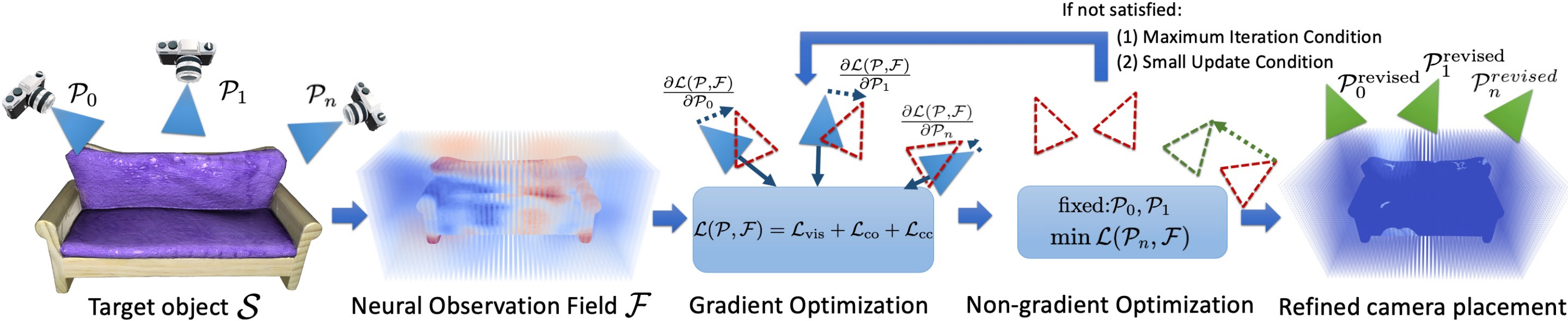}
\vspace{-10mm}
\end{overpic}
\caption{
Method overview. Our method takes target object $\cS$ and initial camera placement $\{ \cP_0, \cP_1,...,\cP_n \}$ as inputs to construct neural observation field $\cF$. We then utilize the non-gradient-based optimization techniques along with gradient-based optimization methods for camera placement refinement. Throughout this optimization process, the neural observation field is continually updated (refer to section ~\ref{algo:hcpo}), until the termination criteria are met.
}
\vspace{-5mm}
\label{fig:pipeline}
\end{figure*}

\subsection{Problem Statement and Method Overview}
Given a target object $\mathcal{S} = \{ \mathbf{s}_j, \bn_j|\mathbf{s} \in \mathbb{R}^3, \bn \in \mathbb{R}^3\}_{j=0:n}$ represented by a point cloud $\{\mathbf{s}\}_{0:n}$ and corresponding normal $\{\bn\}_{0:n}$ (where $n$ indicates the number of points), the camera placement optimization method aims to maximize the visibility of the target scene, such as coverage or observation quality. \changed{The camera placement $\mathcal{P}=\langle \bp_i, \mathbf{r}_i \rangle_{i=1:k}$
involves their positions $\{\mathbf{p}_i | \mathbf{p} \in \mathbb{R}^3\}_{i=1:k}$ and orientations $\{\mathbf{r}_i| \mathbf{r} \in SO(3)\}_{i=1:k}$ (where $k$ indicates the number of cameras)}.
\begin{figure}[t]
\centering
\begin{overpic}
[width=0.95\linewidth]
{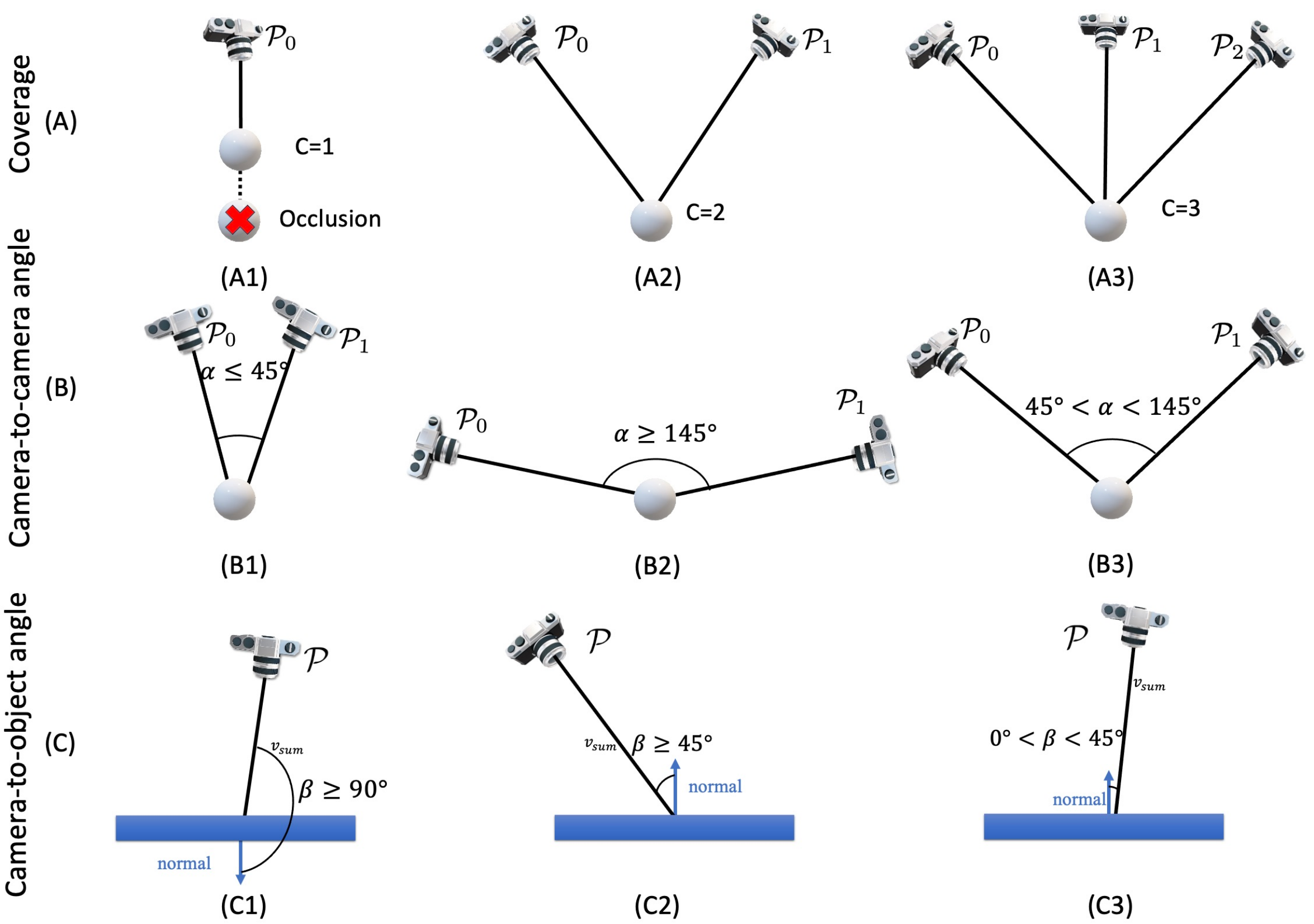}
\end{overpic}
\vspace{-1mm}
\caption{
Illustration of observation attribute $\mathbf{o}$ elements: Coverage $c$ in row (A), Camera-to-camera angle in row (B) and Camera-to-object angle in row (C). Only the third column satisfies our visibility condition.}
\vspace{-6mm}
\label{fig:coverage}
\end{figure}

Our method leverages a hybrid optimization method based on the neural observation field.
~\changed{The neural observation field $\mathcal{F}: (\mathcal{S},\mathcal{P}) \to (c,\bphi^{cc},\bphi^{co})$ implicitly encodes scene priors including \textit{coverage} $(c_i)_{i=1:n}$ (the number of visible cameras), average \textit{camera-to-camera} viewing angle $(\bphi^{cc}_i)_{i=1:n}$ and average \textit{camera-to-object} viewing angle $(\bphi^{co}_i)_{i=1:n}$ (between camera ray and target normal vector)(Section ~\ref{sec:onf}).} Then a hybrid optimization method $\mathcal{H} : (\mathcal{P}_t,\mathcal{S},\mathcal{F}_t)  \to (\mathcal{P}_{t+1},\mathcal{F}_{t+1})$ (Section~\ref{sec:hpo}) is conducted by iteratively performing the gradient-based optimization $\mathcal{H}_{grad}(\mathcal{P}, \mathcal{F})$ and non-gradient-based optimization $\mathcal{H}_{non\_grad}(\mathcal{P}, \mathcal{F})$ to achieve a good trade-off between exploitation and exploration in camera placement optimization. An overview visualization of our pipeline is presented in Figure~\ref{fig:pipeline}.
\subsection{Neural Observation Field}
\label{sec:onf}
\changed{To obtain the visibility of scenes, we divide the target object $\mathcal{S}$ into voxels $\mathcal{V}$. Given the current camera poses, we can determine the voxels that are visible in a single view. After acquiring the observation attributes $o_\mathcal{V}$ of voxels $\mathcal{V}$, we encode the joint configuration of scene geometry and voxels with observation attributes $o_\mathcal{V}$ to derive the neural observation field $\mathcal{F}_t$ of the target object $\mathcal{S}$ and current camera placement $\mathcal{P}_t$}.
\paragraph{Observation Attributes}
\changed{We define the observation attributes, consisting of three elements: Coverage $c$ defined by the coverage relationship $E$, Camera-to-camera angle $\bphi^{cc}$, and Camera-to-object angle $\bphi^{co}$ to represent the visibility of voxels.} We consider the field of view (FOV), image blur, and occlusion to determine visibility. Voxels within the camera's frustum that are not obscured by other voxels in this view are considered visible to that camera.

\changed{The coverage relationship $E(i,j)$ between a voxel $\mathcal{V}_{j}$ and a camera $C_{i}$ is represented as a binary number, where $E(i,j)=1$ if $\mathcal{V}{j}$ is visible in $C_{i}$, otherwise $\changed{E}(i,j)=0$.}
For a given coverage threshold $K$, the coverage condition that voxel $j$ still needs is expressed as :
\begin{equation}
    \changed{c(j)=K-\sum_{i=0}^k{E(i,j)}}.
\end{equation}
When considering the context of multi-view reconstruction, it is imperative to ensure the accuracy of the computed target 3D location. This accuracy relies on the error propagation characteristics of the reconstruction linear solver. In cases where the rays emanating from the view center towards the target point are either parallel or nearly parallel, a valid solution cannot be obtained~\cite{rahimian2016}. Based on the fulfillment of coverage condition, different target points may have varying numbers of rays intersecting them, necessary to adopt a scale-invariant representation to assess the quality of coverage.
To quantify the observation quality of voxel $j$, we utilize two metrics: Camera-to-camera angle $\bphi^{cc}_j$, an internal variability formulated as:
\begin{equation}\label{eq2}
\bphi^{cc}_j = \left | \frac{\pi }{2}-\frac{1}{C_{\left | A_{j}  \right | }^{2}} \sum_{\substack{a_{1}, a_{2} \in A_j \\ a_{1} \neq a_{2}}} \alpha(a_{1}, a_{2}) \right |, 
\end{equation}
and Camera-to-object angle $\bphi^{co}_j$, an external similarity formulated as :
\begin{equation}\label{eq3}
    \bphi^{co}_j = 1-\frac{\bn_{j}\cdot \sum_{a \in A_j} a }{\left \|\bn_j  \right \|\cdot \left \| \sum_{a \in A_j} a  \right \|  }, 
\end{equation}
\changed{where $A_j$ is the set of vectors from the center of voxel $j$ to the cameras that observe voxel $j$, $\alpha$ is the angle between two vectors, and $C_{\left | A_{j} \right | }^{2}$ is a combinatorial term representing the selection of all pairs of vectors from $A_j$.}
The observation attributes consist of three fundamental elements: $\mathbf{o}=[c,\bphi^{cc},\bphi^{co}]$ illustrated in Figure~\ref{fig:coverage} .
\paragraph{Neural Observation Field}
Each voxel $\mathcal{V}_j$ is characterized by its center position $\mathbf{s}_j$, normal $\mathbf{n}_j$ and observation attributes $[c_j,\bphi_j^{cc},\bphi_j^{co}]$. \changed{The neural observation field $\mathcal{F}$ operates by implicitly encoding scene prior perception and offers a continuous and efficient observation query mechanism for optimization, corresponding to the function \textit{LeanNeOF} in Algorithm. ~\ref{algo:hcpo}.}With a fixed number of cameras, surface points observed by each camera are expected to be parts of objects that are less observed in current camera placement.

We employ the Scaled Dot-Product Attention function, which enables the model to focus more on the parts that are relevant to the current or other contextual information~\cite{vaswani2017attention}.~\changed{This function is denoted as $\mathcal{F}$, which aggregates target object surface information to obtain query point attributes and optimization directions to optimize corresponding cameras}.~\changed{Common methods for aggregating information for 3D point clouds or voxels are trilinear interpolation~\cite{kenwright2015free} or KNN algorithm~\cite{qi2017pointnet}.
However, these methods can only aggregate information from nearby small areas, potentially slowing down camera optimization or getting stuck in locally optimal solutions.  According to the concentration mechanism of Attention~\cite{vaswani2017attention}, we can adaptly learn the appropriate weights of all known voxels with attributes via gradient backpropagation.}

We compute the relative position and normal of voxels on $\mathcal{S}$ centered at $\mathbf{S}_i$ and then process through an MLP layer(using ReLU activation and 32 channels) to serve as input $X$ of queries and keys. Subsequently, $X$ is separately multiplied with query and key weight matrices denoted as 
\begin{equation}
Q=XW^Q,\qquad K=XW^K.
\end{equation}
$W^Q, W^K$ are the weight matrices for queries and keys, respectively.
The observation attribute of target object $\mathcal{S}$ is calculated as:

\begin{equation}\label{eq1}
  \mathbf{o}_\mathcal{S} = softmax(\frac{QK^{T} }{\sqrt{d_{k} } } )\ \mathbf{o}_{\mathcal{V}},
\end{equation}
where $d_k$ is the dimension of the keys. Calculating by Eq.~\ref{eq1}, ~\changed{target object $\mathcal{S}$ have equivalent attributes $\mathbf{o}_\mathcal{S}$.}
 According to Eq.\ref{eq2}, \ref{eq3}, \changed{Attribute $\mathbf{o}_\mathcal{S}=[c,\bphi^{cc}\bphi^{co}]$ has a maximum value $sup=[K,\pi/2,1]$}. The sum attributes of all points in all views have a maximum value as well :
\begin{equation}\label{eqsup}
    \changed{\sum_{i=1}^{k}\sum_{j=1}^{\left | \mathcal{S}_i \right | }\mathbf{o}_{\changed{\mathcal{S}}_{ij}} \le   \sum_{i=1}^{k} \sum_{j=1}^{n}sup}, 
\end{equation}
where $\changed{\left | \mathcal{S}_i \right |}$ is the number of points in view $i$~\changed{and $\mathcal{S}_{ij}$ is the $jth$ point of points in view $i$}. 
~\changed{After generating the Neural Observation Field, we optimize the camera placement by a hybrid optimization method.}

\IncMargin{0.5em}
\begin{algorithm}[t]
\caption{Hybrid camera placement optimization $\mathcal{H}$}
\label{algo:hcpo}
\SetKwInOut{AlgoInput}{Input}
\SetKwInOut{AlgoOutput}{Output}
\SetKwFunction{ShapeAnalyze}{ShapeAnalyze}
\SetKwFunction{LeanNeOF}{LeanNeOF}
\SetKwFunction{GradientOptimize}{GradientOptimize}
\SetKwFunction{TrustRegionOptimize}{TrustRegOptimize}
\SetKwFunction{CheckLossUpdate}{CheckLossUpdate}
\SetKwFunction{CheckStepUpdate}{CheckStepUpdate}
\SetKwFunction{Init}{Initialize}
\AlgoInput{~\changed{Target object $\mathcal{S}$}}
\AlgoOutput{ Optimized camera placement $\mathcal{P}$}
$\mathcal{P}_0 \leftarrow$ \textit{Initialize}($\mathcal{S}$)\; 
$c_0, \phi^{co}_0, \phi^{cc}_0 \leftarrow$ \textit{ShapeAnalyze}($\cS$, $\cP_0$)\;
$\cF_0 \leftarrow$ \textit{LeanNeOF}($c_0, \phi^{co}_0, \phi^{cc}_0, \cS $ )\;
$\cL_0 \leftarrow \infty$;  $t \leftarrow 0$; \tcp*[f]{\small Initialize the NeOF}  \\ex

\Repeat{\textit{CheckStepUpdate}($\cP_t$, $\cP_{t-1}$) $< 10^{-4}$ }{ 
    
    $\cP_{t+1}, \cL_{t+1} \leftarrow$ \textit{$\mathcal{H}_{grad}$}($\cP_t, \cF_t$)\;
    $c_{t+1}, \phi^{co}_{t+1}, \phi^{cc}_{t+1} \leftarrow$ \textit{ShapeAnalyze}($\cS$, $\cP_{t+1}$)\;
    $\cF_{t+1} \leftarrow$ \textit{LeanNeOF}($c_{t+1}$, $\phi^{co}_{t+1}$, $\phi^{cc}_{t+1}$, $\cS$, $\cF_t$ )\; 
    \tcp*[f]{\small Fine-tune the NeOF}  \\
    \If {\textit{CheckLossUpdate}($\cL_{t}, \cL_{t+1}$)$<1e^{-4}$}{  
        $\cP_{t+1} \leftarrow$ \textit{$\mathcal{H}_{non\_grad}$}($\cP_{t+1}, \cF_{t+1}$) \;
        $c_{t+1}, \phi^{co}_{t+1}, \phi^{cc}_{t+1} \leftarrow$ \textit{ShapeAnalyze}($\cS$, $\cP_{t+1}$)\;
        $\cF_{t+1} \leftarrow$ \textit{LeanNeOF}($c_{t+1}, \phi^{co}_{t+1}, \phi^{cc}_{t+1}, \cS, \cF_{t+1}$ )\;
            \tcp*[f]{\small Extensively fine-tune the NeOF}  \\
    }

    $\cL_{t} \leftarrow \cL_{t+1}$; $t \leftarrow t+1$\;

}
\end{algorithm}
\vspace{-8pt}
\subsection{\textbf{Hybrid Placement Optimization method}}
\label{sec:hpo}
Utilizing the differentiable and efficient scene priors query facilitated by the neural observation field, our approach strategically employs both gradient-based optimization and non-gradient-based optimization to strike a favorable balance between exploitation and exploration. Gradient-based optimization enables direct access to gradients derived from the neural observation field, facilitating fine-grained optimization. Nevertheless, it is susceptible to being trapped in local optima.~\changed{To mitigate this risk and escape local optima, non-gradient-based optimization comes into play. Non-gradient-based optimization identifies and stabilizes camera poses associated with superior coverage and observation quality, subsequently reevaluating sub-optimal camera poses by leveraging analysis of scene priors and convergent results of gradient-based optimization method.} For a more comprehensive understanding of the methodology, the details can be found in Algorithm. ~\ref{algo:hcpo}.

\textbf{Gradient-based camera placement optimization.}
Gradient optimization is widely used to have fine-grained updating steps for high-quality optimization. With the differentiable neural observation field $\cF$, our method can obtain the gradient via a self-defined loss. 
~\changed{Based on Eq. \ref{eq1}, \ref{eqsup}, maximizing the coverage and observation quality is equivalent to minimize
\begin{equation}\label{eqloss}
[\cL_{vis},\cL_{cc},\cL_{co}] = sup-\frac{\sum_{i=1}^{k}\sum_{j=1}^\changed{{\left | \mathcal{S}_i \right | }}\mathbf{o}_{\changed{\mathcal{S}}_{ij}} }{k\cdot n}, 
\end{equation}
where $[\cL_{vis},\cL_{cc},\cL_{co}]$ are three different metrics, including (1) Coverage loss $\cL_{vis}$ guiding the camera to maximize its observation of objects, (2) Camera-to-camera viewing angle loss $\cL_{cc}$, to enable triangular perception, and (3) Camera-to-object viewing angle loss $\cL_{co}$, to enable cameras to face directly to the object.} 

\changed{To increase the coverage and observation quality, we use $\mathcal{H}_{grad}(\mathcal{P}, \mathcal{F})$ to optimize the camera placement $\mathcal{P}=\langle \bp_i, \br_i \rangle_{i=1:k}$ through a combination loss of three metrics:}
\begin{eqnarray}\label{equ:loss}
    \begin{split}
    \cL(\cP,\cF) = w_{vis}\cL_{vis} + w_{cc}\cL_{cc} + w_{co}\cL_{co},\\
    \end{split}
\end{eqnarray}
where the $w_{vis}$, $w_{cc}$ and $w_{co}$ are the weights of each cost term. We empirically set the weights as $w_{vis}=0.4$, $w_{cc}=0.3$ and $w_{co}=0.3$, and the weights can be flexibly adjusted for specific requirements, like maximizing the visibility ($w_{vis}=1.0$, $w_{cc}=0.0$ and $w_{co}=0.0$). 

\changed{Camera poses $\mathcal{P} = \langle \mathbf{p}, \mathbf{r}\rangle$ are optimized using the gradient of $\cL(\cP, \cF)$ as the optimization parameters. The camera transforms the point cloud observed in its view to the Neural Observation Field coordinate system using its camera parameters and calculates the loss. Due to this differentiable forward propagation, the gradient can backpropagate to the camera parameters using \textit{PyTorch}~\cite{paszke2017automatic}.}

\changed{After each iteration of gradient-based optimization, our method recalculates the visibility, as \textit{ShapeAnalyze} and fine-tunes the neural observation field with significant down-sampled points for efficiency, as \textit{LeanNeOF}. The gradient-based optimization stops when the gradient of camera parameters and the difference between two consecutive losses is less than 1e-4. This stopping criterion corresponds to \textit{CheckStepUpdate} in Algorithm. ~\ref{algo:hcpo}.} 


\textbf{Non-gradient-based camera placement optimization.}
\changed{The gradient optimization performs well when the object exhibits weak non-convexity. However, objects in real-world scenarios are often diverse and highly non-convex. To escape from local optima, we employ a non-gradient-based optimization method $\mathcal{H}_{non\_grad}(\mathcal{P}, \mathcal{F})$, akin to trust-region optimization. This method fixes the well-optimized camera placements and recalculates the positions and orientations of less optimal cameras.}

\changed{We update the camera that satisfies two conditions: first, the gradient is less than 1e-4, indicating that the camera has converged; second, the camera still has a large loss after convergence, indicating a better observation camera pose in global space than this one.} 
To replace this camera, we filter the $m$ least-covered regions to calculate the current camera placement neural field. The coverage and observation quality are compared using Equation \ref{equ:loss} to generate a new camera pose, set directly above the poorly covered area of the object surface.
After generating a new camera pose, we update the neural observation field. After traversing throughout $m$ regions, we choose the best camera pose with the minimum $\cL$:
\changed{\begin{equation}
     p_{w_{new}}=\mathop{\arg\min}\limits_{j }\ \cL(\cP_{t_{i\to j}},\cF_t)_{j=1:m}.
\end{equation}}
We continue replacing worse camera poses until no better new camera pose can be found.
\subsection{Implementation details}
The initialization of camera placements is randomly sampled points within the space of the scene. For differentiable optimization, we use the Adam optimizer to optimize parameters containing both camera poses and the Attention layer. The initial learning rate of the optimizer is $1e^{-3}$, with a gradual decrease during the optimization process.
While dealing with occlusion in view, we employ spherical inverse flipping and convex hull construction from \cite{katz2007direct} to obtain points not hidden by other points in one view, reducing the rendering time and input complexity.

\section{Experiments}
\begin{figure}[t]
\centering
\begin{overpic}
[width=\linewidth]
{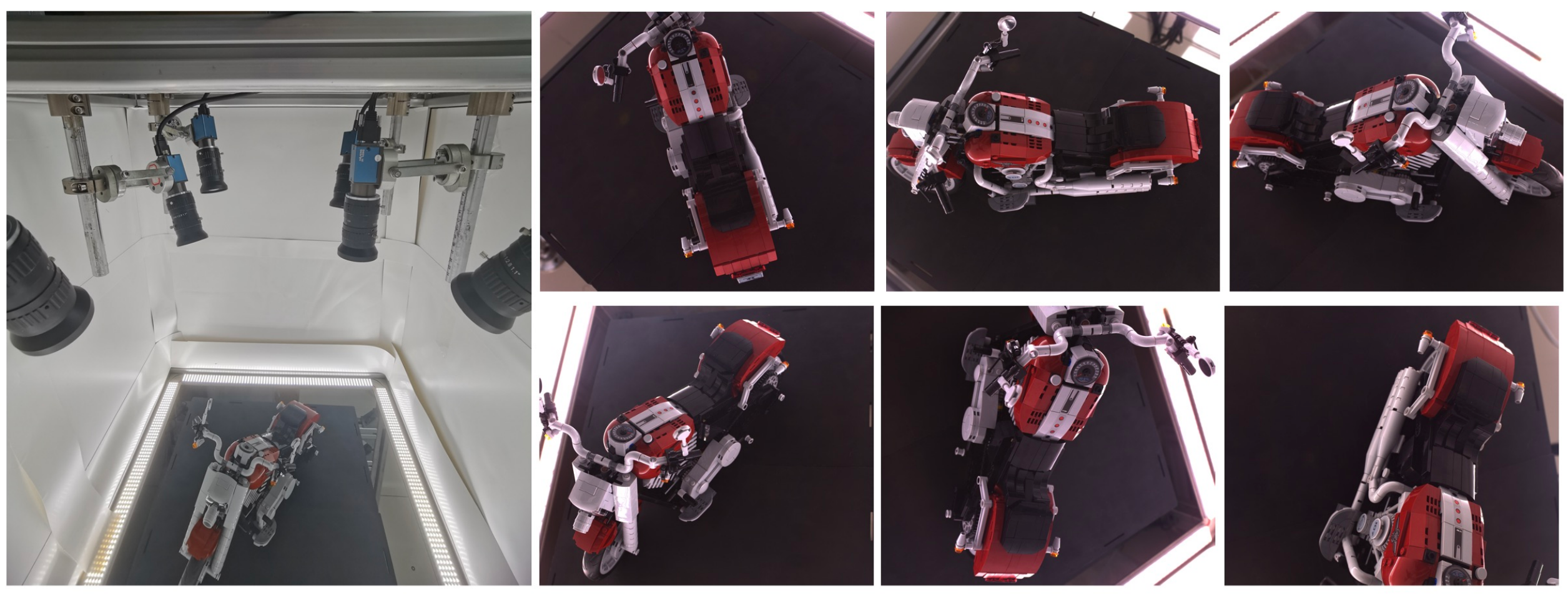}
\end{overpic}
\caption{
Camera placement control system. This system is able to control six 4K RGB cameras in SE(3) with a uniform distributed lighting source. Real images captured by optimized cameras.
}
\vspace{-7mm}
\label{fig:real_world_system}
\end{figure}

\subsection{Experiment setup}


\textbf{Synthetic environments setup:}. The synthetic environment is composed of 2D planar shapes (25), 3D objects (28), and room-scale 3D scenes (8), leading to a total number of 56 which is more than existing methods.
The 2D planar shapes are self-generated with random shape (circle, triangular, cube) combinations.
For the 3D object models and 3D scenes, we leverage high-quality real scanned reconstruction from Google Scanned Objects (GSO) dataset~\cite{Downs2022GoogleSO} and Replica dataset~\cite{straub2019replica}, respectively.

\textbf{Real-world experiments setup:} We construct a camera placement control system as shown in Fig.~\ref{fig:real_world_system}, composed of six 4K RGB cameras and constant lighting sources. All six cameras can be freely adjusted to capture RGB images. The camera placement can be dynamically adjusted based on the results of camera placement optimization methods. Here, we leverage five diverse objects including the bag, flower, laptop, and motor Lego, for evaluating the performance of methods in the real-world environment.

\begin{table*}[t]\centering
\caption{
Comparing Coverage optimality gap and Observation angle quality in 2D Plane and 3D Model datasets. We present the average outcomes for each method across varying numbers of cameras. The best results are highlighted. Here, cam. indicates the number of cameras.
}
\vspace{-2mm}
\scalebox{0.95}{
\setlength{\tabcolsep}{0.8mm}
\renewcommand{\arraystretch}{1.2}
\begin{tabular}{l|cccc|cccc|cccc|cccc}
\hline
\multirow{2}{*}{Methods} & \multicolumn{8}{c|}{2D planar shapes} & \multicolumn{8}{c}{3D objects} \\ \cline{2-17} 
& \multicolumn{4}{c|}{Coverage optimality gap $\downarrow$} & \multicolumn{4}{c|}{Observation angle quality $\uparrow$} & \multicolumn{4}{c|}{Coverage optimality gap $\downarrow$} & \multicolumn{4}{c}{Observation angle quality $\uparrow$} \\ \hline \hline
Init & 0.77 & 0.54 & 0.37 & 0.28 & 0 & 0 & 0.44 & 0.69 & 0.75 & 0.51 & 0.34 & 0.29 & 0.39 & 0.93 & 0.91 & 0.80 \\
DE\cite{rangel2019} & 0.75 & 0.64 & 0.52 & 0.50 & 0.09 & 0.10 & 0.58 & 0.64 & 0.82 & 0.67 & 0.56 & 0.46 & 0.34 & 0.91 & 0.86 & 0.88 \\
PSO\cite{jiao2019coverage} & 0.77 & 0.61 & 0.52 & 0.45 & 0.02 & 0.28 & 0.54 & 0.52 & 0.83 & 0.72 & 0.31 & 0.27 & 0.49 & 0.88 & 0.85 & 0.84 \\
GA \cite{Navin2010SolvingCP} & 0.87 & 0.79 & 0.71 & 0.70 & 0.00 & 0.09 & 0.37 & 0.37 & 0.83 & 0.76 & 0.37 & 0.34 & 0.41 & 0.80 & 0.85 & 0.86 \\
SA\cite{shahrokhzadeh2017} & 0.77 & 0.55 & 0.36 & 0.28 & 0.03 & 0.15 & 0.54 & 0.67 & 0.76 & 0.54 & 0.31 & 0.24 & 0.45 & 0.79 & 0.81 & 0.69 \\
MIP\cite{malhotra2022} & 0.68 & 0.58 & 0.55 & 0.53 & 0.53 & 0.45 & 0.40 & 0.38 & 0.69 & 0.52 & 0.45 & 0.38 & 0.62 & 0.83 & 0.82 & 0.81 \\
Ours & \textbf{0.64} & \textbf{0.43} & \textbf{0.30} & \textbf{0.17} & \textbf{0.57} & \textbf{0.54} & \textbf{0.73} & \textbf{0.72} & \textbf{0.63} & \textbf{0.40} & \textbf{0.28} & \textbf{0.20} & \textbf{0.70} & \textbf{0.96} & \textbf{0.91} & \textbf{0.89} \\ \hline
\end{tabular}
}
\vspace{-2mm}
\label{tab:benchmark}
\end{table*}

\textbf{Baseline methods.} 
Coverage in multi-media placement is an NP-hard problem \cite{bouhali2021}, and various approaches have been proposed to solve this problem. We compare our methods with mainstream methods:
\begin{itemize}
    \item \textbf{Genetic Algorithm (GA) }in \cite{Navin2010SolvingCP}, emulates natural process to select, generate and evaluate for fitness.
    \item \textbf{Simulated Annealing (SA) }in \cite{shahrokhzadeh2017}, iteratively replaces inferior solutions until algorithm's temperature reaches a predefined threshold, akin to metal heat treatment.
    \item \textbf{Particle Swarm Optimization (PSO) } in \cite{puligandla2023continuous}, leverages the exchange of information among individuals within a group to facilitate the transition.
    \item \textbf{Differential Evolution (DE) }in \cite{rangel2019}, distinguishes from GA by crossing with parent individual vectors to generate new ones, proving more effective than GA.
    \item \textbf{Mixed-Integer Programming (MIP) }in \cite{malhotra2022}, employs a branch-and-bound algorithm, solving a sequence of linear programming derived from original problem.
    \item \textbf{Graph Neural Network (GNN) }in \cite{zhou2022} leverages a graph neural network to capture local interactions of the robots on 2D planar shapes.
\end{itemize}

\begin{table}[!t]\centering
\caption{
Comparing Coverage optimality gap and Observation angle quality in 3D Scene datasets, same as tested in 2D Plane and 3D Model datasets.}
\vspace{-2pt}
\scalebox{0.9}{
\setlength{\tabcolsep}{2mm}{
\begin{tabular}{lcccccc}
\hline
\multicolumn{1}{l|}{\multirow{2}{*}{Method}} & \multicolumn{3}{c|}{Coverage optimality gap $\downarrow$}                                               & \multicolumn{3}{c}{Observation angle quality $\uparrow$}                             \\ \cline{2-7} 
\multicolumn{1}{l|}{}                        & \multicolumn{1}{c|}{10 cam.} & \multicolumn{1}{c|}{20 cam.} & \multicolumn{1}{c|}{30 cam.} & \multicolumn{1}{c|}{10 cam.} & \multicolumn{1}{c|}{20 cam.} & 30 cam.  \\ \hline \hline
Init. & 0.23 & 0.12 & 0.09 & 0.07 & 0.07 & 0.06 \\
DE~\cite{rangel2019} & 0.20 & 0.08 & 0.04 & 0.31 & 0.26 & 0.28 \\
GA~\cite{Navin2010SolvingCP} & 0.38 & 0.13 & 0.06 & 0.33 & 0.37 & 0.38 \\
PSO~\cite{jiao2019coverage} & 0.21 & 0.08 & 0.05 & 0.55 & 0.45 & 0.44 \\
SA~\cite{shahrokhzadeh2017} & 0.20 & 0.07 & 0.04 & 0.44 & \textbf{0.47} & 0.42 \\ \hline
Ours & \textbf{0.16} &\textbf{0.04} &\textbf{0.03} & \textbf{0.58} & 0.46 & \textbf{0.47} \\ \hline
\end{tabular}
}}
\vspace{-1pt}
\label{tab:3dscenes}
\end{table} 
\begin{figure}[t]
\centering
\begin{overpic}
[width=\linewidth]
{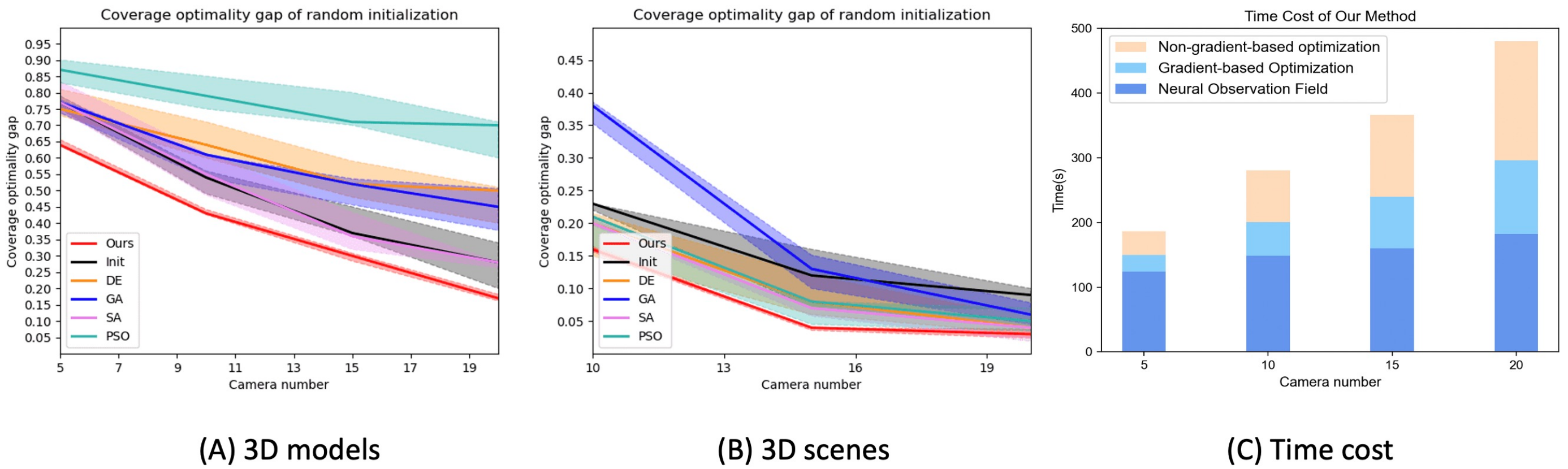}
\end{overpic}
\caption{
The robustness and time cost of our algorithm, we optimize 10 sets of camera placements with different initializations. The solid line is the mean value, with the shading represents the upper and lower bounds. Additionally, we have tested the time cost of different parts in our method. 
}
\vspace{-16pt}
\label{fig:curve}
\end{figure}

We compare results with GA, SA, PSO, DE, MIP methods on all 2D planar shapes, 3D objects and room-scale 3D scenes. For GNN, we test only on 2D planar shapes due to the limitation of their method.  

\textbf{Metrics.} 
We plan to evaluate the camera placement in terms of both coverage and observation quality.
For coverage evaluation,we consider \textbf{Coverage optimality gap}~\cite{malhotra2022}, given by the formula: 
\begin{equation}
    uc=\frac{(K-\sum_{i=0}^{n}cov_i)^2}{K\cdot n^2}, 
\end{equation}
where $K$ represents the required coverage number, (we use 3 in our experiment),
$n$ is the number of total visible points, and $cov_i$ represents the number of cameras able to observe point $i$ in current camera placement. Smaller values of the Coverage optimality gap indicate better performance.

\begin{figure}[t]
\centering
\begin{overpic}
[width=0.8\linewidth,height=0.69\linewidth]
{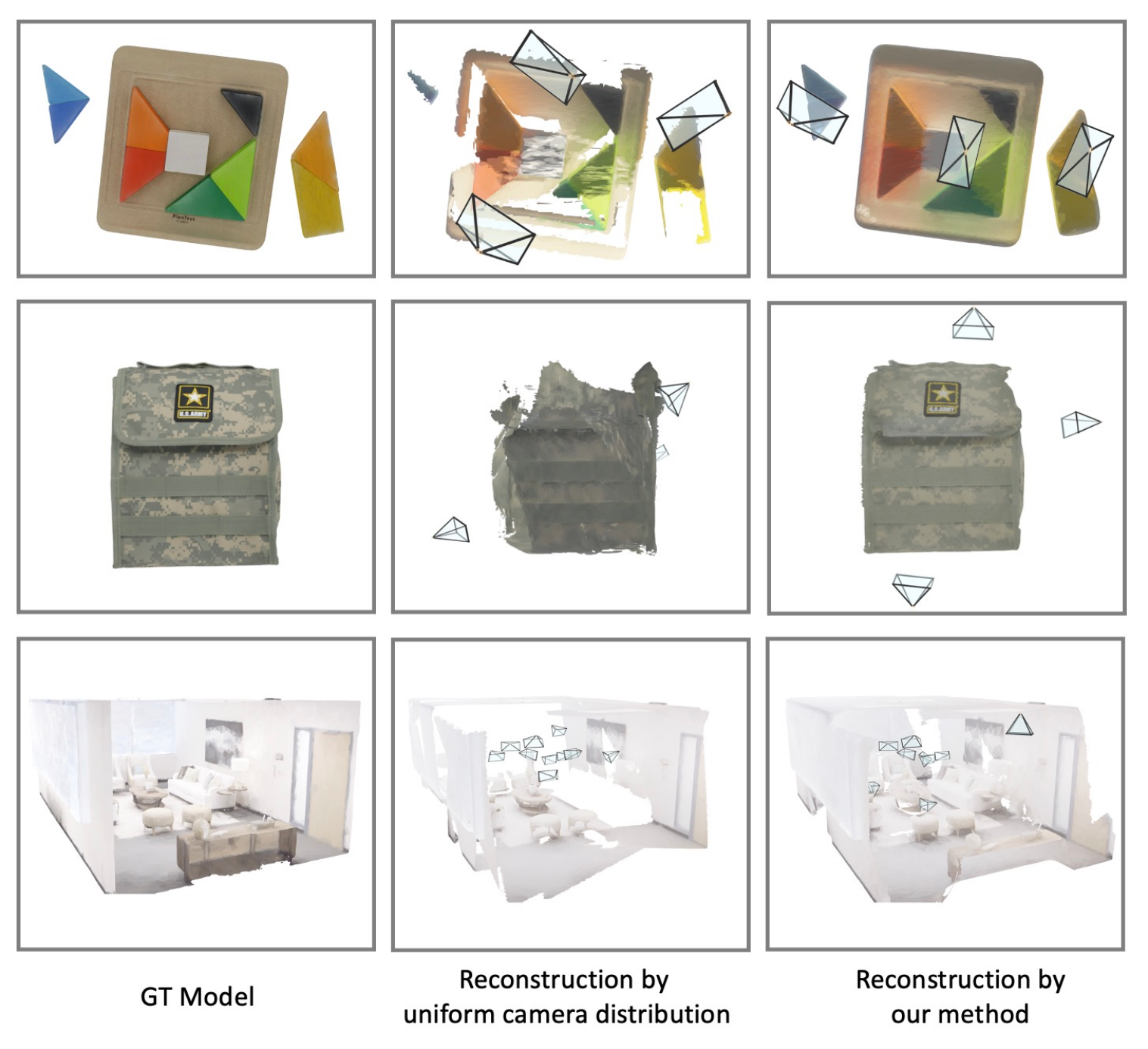}
\end{overpic}
\vspace{-2mm}
\caption{
Comparisons of reconstruction completeness between 3 cameras uniformly distributed and by our optimization method. Missing parts are ones not observed by any camera. 
}
\vspace{-14pt}
\label{fig:reconstruction}
\end{figure}

\begin{figure*}[t]
\centering
\begin{overpic}
[width=0.8\linewidth,height=0.40\linewidth]
{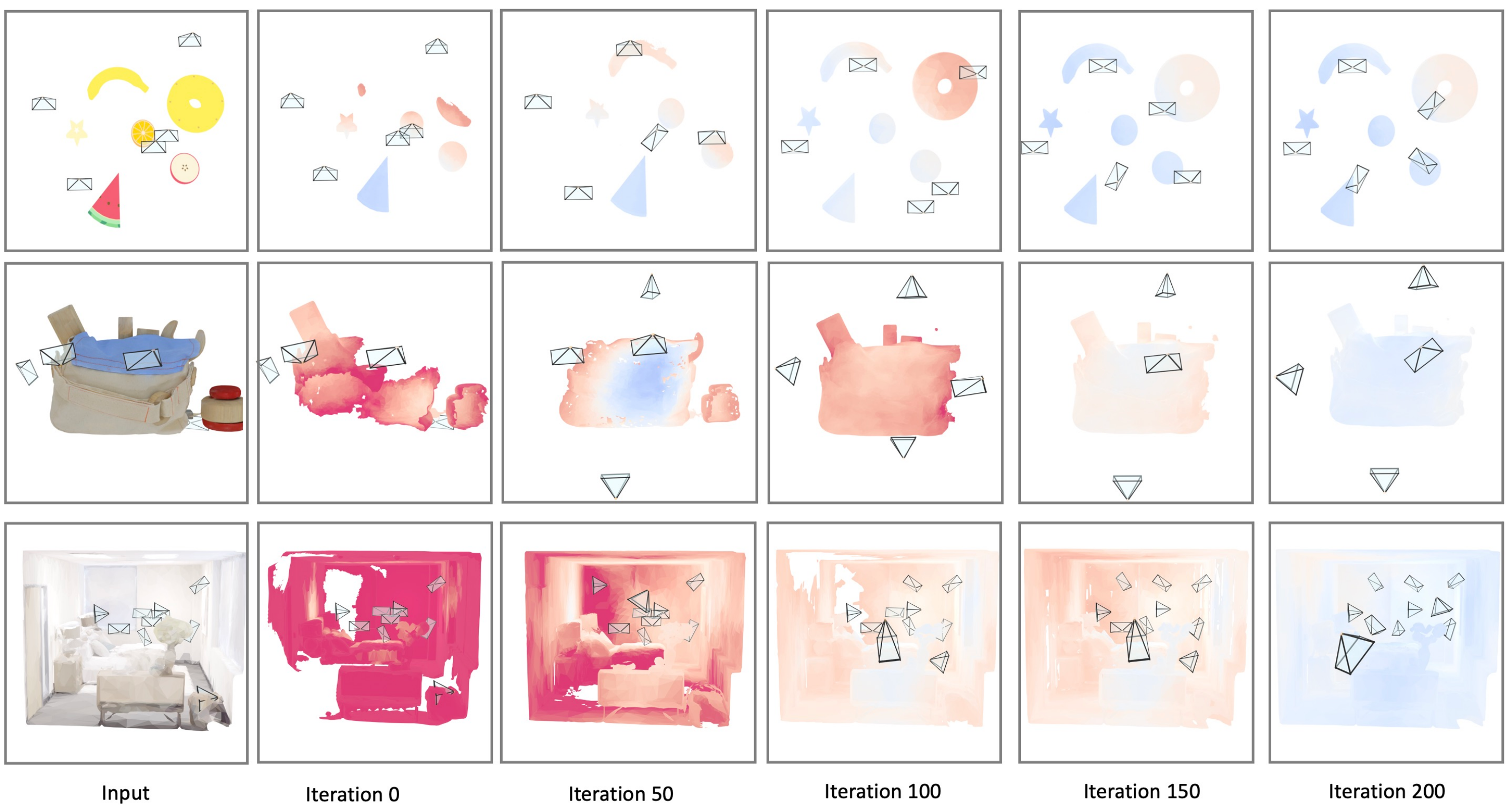}
\end{overpic}
\vspace{-2pt}
\caption{
Visual results of neural observation field during optimization. From left to right, the optimization step increases and the color to blue, the better observation and reconstruction quality. 
}
\vspace{-14pt}
\label{fig:visual}
\end{figure*}

For quality evaluation, we consider the Camera-to-object angle and the Camera-to-camera angle as proposed in Section \ref{sec:onf}. The angles among vectors from camera rays to the target point within $[45^\circ,145^\circ]$ has been shown to approximate triangular perception, leading to better observation quality~\cite{rahimian2016}. 
We calculate the rate of satisfied rays among all camera-to-camera rays as an angle rate to assess the quality, called \textbf{Observation angel quality}.
\subsection{Results and analysis}
\textbf{Comparison on 2D planar shapes, 3D objects, Room-scale 3D scenes.} 
We extensively compare our methods in diverse environments (2D planar shapes, 3D objects, and 3D scenes) with different camera numbers. 
The results of 2D planar shapes and 3D object datasets can be found in Table~\ref{tab:benchmark}, and for 3D scenes, the results are presented in Table~\ref{tab:3dscenes}.
~\changed{We use an acceptable number of particles (30) to test all datasets. We also perform extensive experiments with different particle numbers, finding that the methods reach convergence as the particle number increases. While these methods are comparable with ours in terms of performance when converged, their time overhead is at least one hour, which is 48 times longer than ours.}
The results prove that our method has significant advantages in both coverage and observation quality over existing methods across different camera numbers. Specifically, we find our method showcases stable improvement along with the increase in the number of cameras, constantly showing better performance than existing methods. As for 3D scene datasets, we report the results in Table~\ref{tab:3dscenes}, where we continue to achieve state-of-the-art performance compared to all methods. Experiments on synthetic datasets show the applicability of our approach in various scenarios. The corresponding visual results can be found in Fig.~\ref{fig:visual}, where models from crippled or red to complete and blue, reflecting our optimization process. To exclude the effect of initialization, we perform 10 times experiments with different random initializations to verify the robustness of the approach in Fig.~\ref{fig:curve}. Thin shading bounds of our method illustrate its non-dependency on the quality of initial camera placement. The visualization of reconstruction shown in Fig.~\ref{fig:reconstruction} as a downstream task, illustrates better completeness and accuracy of reconstructed models achieved by our optimization method. 

\begin{table}[!t]\centering
\caption{
Comparing single-coverage of learning-based GNN method~\cite{zhou2022} with our method.We compare results on their self-generated plane dataset with single-coverage metric, the percentage of covered targets number. Here, cam. indicates the number of cameras.}
\vspace{-2pt}
\scalebox{0.9}{
\setlength{\tabcolsep}{4mm}{
\begin{tabular}{lcccc}
\hline
\multicolumn{1}{l|}{\multirow{2}{*}{Method}} & \multicolumn{4}{c}{Single-coverage metric~\cite{zhou2022} $\downarrow$}
\\ \cline{2-5} 
\multicolumn{1}{l|}{}  & \multicolumn{1}{c|}{10 cam.} & \multicolumn{1}{c|}{20 cam.} & \multicolumn{1}{c|}{30 cam.} & 40 cam.  \\ \hline \hline

GNN \cite{zhou2022}                          & 0.41                     & 0.36                     & 0.23                     & 0.14 \\
Ours                         & \textbf{0.36}                     & \textbf{0.33}                     & \textbf{0.21}                     & \textbf{0.13} \\ \hline
\end{tabular}
}}
\vspace{-6mm}
\label{tab:comp_learning}
\end{table} 
\textbf{Comparison with the learning-based method~\cite{zhou2022}.}
In addition to traditional methods, we experiment to compare with the learning-based method presented in \cite{zhou2022}. Camera poses in \cite{zhou2022} can only optimize on 2D planar shapes, thus we adjust ours accordingly. 
The results are provided in Table~\ref{tab:comp_learning}.
Our method outperforms the GNN method~\cite{zhou2022} for various camera placement numbers, which is the state-of-the-art camera placement optimization on 2D planar shapes. 
However, it's worth noting that our advantage diminishes as the number of cameras increases because it becomes easier to achieve better coverage with more cameras.  

\begin{table}[!t]\centering
\caption{
Ablation study of Hybrid optimization and neural observation field. Rows 1-2 use only gradient-based or non-gradient-based optimization methods, and rows 3-5 use discrete or distance-based neural fields as comparison. 
}
\vspace{-1pt}
\scalebox{0.75}{
\setlength{\tabcolsep}{1mm}{
\begin{tabular}{lcc}
\hline
\multicolumn{1}{l|}{Method} & \multicolumn{1}{c|}{Coverage optimality gap $\downarrow$} & Observation angle quality $\uparrow$ \\ \hline \hline
Grad. opt. + neural obs. filed      & 0.510 & 0.880 \\
\changed{Non-grad. opt. + neural obs. field} & \changed{0.473} & \changed{0.896} \\
Hybrid opt. + trilinear interpolation     & 0.588 & 0.818 \\
Hybrid opt. + neural distance field   & 0.512 & \textbf{0.937} \\
Hybrid opt. + neural obs. field (Ours)      & \textbf{0.398} & 0.933 \\ \hline
\end{tabular}
}
}
\vspace{-2pt}
\label{tab:ablation}
\end{table} 
\begin{table}[!t]\centering
\caption{
Real-world experiments on our camera placement control system, we dynamically optimize the placement of six cameras on five real-world models and obtain the coverage and observation quality. }
\vspace{-2pt}
\scalebox{0.78}{
\setlength{\tabcolsep}{6mm}{
\begin{tabular}{lcc}
\hline
\multicolumn{1}{l|}{Models} & \multicolumn{1}{c|}{Coverage optimality gap $\downarrow$} & Observation angle quality$\uparrow$ \\ \hline \hline
Moto                       & 0.06                              & 0.87          \\
Laptop                      & 0.00                              & 0.81          \\
Flower                      & 0.10                              & 0.72          \\
Sandbox                     & 0.00                              & 0.92          \\
Bag                         & 0.03                              & 0.88          \\ \hline
\end{tabular}
}}
\vspace{-3mm}
\label{tab:realworld}
\end{table} 
\textbf{Ablation Study.}
We conduct an ablation study to verify the effectiveness of key components in our method using 2D planar shapes dataset. 
\changed{Optimization targets of all methods are both  Coverage optimality gap and Observation angle quality.}
The results can be found in Table~\ref{tab:ablation}, demonstrates better performance than gradient- or non-gradient optimization (rows 1-2) as the gradient optimization method suffers from a highly non-linearity optimization landscape. While combining it with non-gradient optimization, we observe a significant improvement which proves the effectiveness of using hybrid optimization. Moreover, rows 3-5 demonstrate that our neural observation field outperforms other alternatives, underscoring its capability to provide both differentiable and observation measurements to guide the optimization, resulting in better observation quality that underlies coverage.

\textbf{Real-world experiments.}
The real-world experiments conducted using our camera placement control system are summarized in  table~\ref{tab:realworld}. Here, we leverage five real-world models and dynamically adjust the camera placement by our optimization, with input shapes scanned roughly. The results clearly demonstrate that our method still has good performance in real-world environments by almost fully covering the target objects while having good camera quality, capturing information from concave planes of objects as well.

\textbf{Analysis of Computational Cost.} 
Our method exhibits high efficiency in both learning the neural observation field and the optimization process shown in Fig.~\ref{fig:curve}. Under typical 10-camera seniors, our method requires only 0.08 seconds per iteration for learning the neural observation field and 0.1 seconds for the optimization process in each iteration. Throughout our experiments, our method consistently exhibited the fastest speed (eight times faster than mainstream methods, as reported in~\cite{malhotra2022,puligandla2023continuous,rangel2019,jiao2019coverage}), while also achieving state-of-the-art performance.

\section{Conclusion}

In this work, we present a novel camera placement hybrid optimization method leveraging the neural observation field. Our method amalgamates the strengths of both gradient-based and non-gradient-based optimization techniques, striking a balance between their respective advantages. To enable a unified observation for both methods, the neural observation field learns the coverage and observation quality of camera placement in a differentiable manner. The results on both synthetic datasets and real-world datasets clearly demonstrate the superiority of our methods. In the future, we would like to exploit
our approach under dynamic environments or large-scale scenes \textit{e.g.} a whole building.

\bibliographystyle{IEEEtran}
\bibliography{IEEEabrv,reference}







\end{document}